\documentclass[conference]{IEEEtran}
\IEEEoverridecommandlockouts
\usepackage{cite}
\def\BibTeX{{\rm B\kern-.05em{\sc i\kern-.025em b}\kern-.08em
    T\kern-.1667em\lower.7ex\hbox{E}\kern-.125emX}}
    
\usepackage{amsmath,amsfonts, amsthm, bm, amssymb}

\def\Figref#1{Figure~\ref{#1}}

\def\eqref#1{equation~(\ref{#1})}
\def\plainref#1{(\ref{#1})}

\def\1{\bm{1}}

\def\rk{{\textnormal{k}}}

\def\vzero{{\bm{0}}}

\def\vtheta{{\bm{\theta}}}

\def\vphi{{\bm{\phi}}}

\def\ve{{\bm{e}}}

\def\vx{{\bm{x}}}
\def\vy{{\bm{y}}}

\def\mA{{\bm{A}}}

\def\mD{{\bm{D}}}

\def\mW{{\bm{W}}}
\def\mX{{\bm{X}}}

\def\mPhi{{\bm{\Phi}}}

\def\mSigma{{\bm{\Sigma}}}

\DeclareMathAlphabet{\mathsfit}{\encodingdefault}{\sfdefault}{m}{sl}
\SetMathAlphabet{\mathsfit}{bold}{\encodingdefault}{\sfdefault}{bx}{n}

\def\gO{{\mathcal{O}}}

\newcommand{\R}{\mathbb{R}}

\DeclareMathOperator*{\argmin}{arg\,min}

\def\<#1,#2>{\langle #1,\,#2\rangle}

\usepackage{url}
\usepackage{xcolor}
\usepackage[colorlinks = true,
            linkcolor = blue,
            urlcolor  = blue,
            citecolor = blue,
            anchorcolor = blue]{hyperref}

\usepackage{enumitem}
\usepackage{rotating}
\usepackage{tabulary}
\usepackage{multirow}
\usepackage[ruled, lined]{algorithm2e} % ,linesnumbered

\usepackage{graphicx, epstopdf}
\usepackage{wrapfig}
\usepackage{caption}
\usepackage{subcaption}
\graphicspath{{./figs/}}

\newtheorem{theorem}{Theorem}

\newtheorem{proposition}{Proposition}

\usepackage{booktabs} % for professional tables    

\begin{document}
\title{NNK-Means: Data summarization using dictionary learning with non-negative kernel regression \\
\thanks{Our work was supported in part by a DARPA grant (FA8750-19-2-1005) in Learning with Less Labels (LwLL) program and by NSF (CCF-2009032).}
}

\author{\IEEEauthorblockN{Sarath Shekkizhar, Antonio Ortega}
\IEEEauthorblockA{
\textit{University of Southern California}\\
Los Angeles, CA, USA \\
\{shekkizh, aortega\}@usc.edu}
}

\maketitle

\begin{abstract}
An increasing number of systems are being designed by gathering significant amounts of data and then optimizing the system parameters directly using the obtained data. 
Often this is done without analyzing the  
dataset structure.  
As task complexity, data size, and parameters all increase to millions or even billions, {\em data summarization} is becoming a major challenge.
In this work, we investigate data summarization via 
 dictionary learning~(DL),
leveraging the properties of recently introduced non-negative kernel regression (NNK) graphs. Our proposed NNK-Means, unlike previous DL techniques, such as kSVD, learns geometric dictionaries with atoms that are representative of the input data space. Experiments show that summarization using NNK-Means can provide better class separation compared to linear and kernel versions of kMeans and kSVD.  Moreover, NNK-Means is scalable, with runtime complexity similar to that of kMeans.
\end{abstract}


\section{Introduction}
\label{sec:introduction}
Massive high-dimensional datasets are becoming an increasingly common input for system design. 
While large datasets are easier to collect, the methods for exploratory (understanding or characterizing the data) and confirmatory (confirming the validity and stability of a system designed using the data) analysis are not as scalable and require new techniques that can cope with big data sizes~\cite{tukey1977exploratory, jain2010data}.    
Data summarization methods aim to represent large datasets by a small set of elements, the insights from which can be used to organize the dataset into clusters, classify observations to its clusters, or detect outliers \cite{leskovec2020mining}.
In datasets with label information, a summary can be obtained for each class, but summaries are in general decoupled from downstream data-driven system designs 
% and learning algorithms 
and thus different from  coresets and sketches~\cite{phillips2016coresets,munteanu2018coresets}.

Clustering methods such as kMeans \cite{lloyd1982least, jain2010data}, vector quantization \cite{gray1984vector} and their variants \cite{kleindessner2019fair}, are among the most prevalent approaches to data summarization \cite{gan2020data}. 
% These methods have been used to solve various data analysis problems by leveraging substructures in the dataset and reasoning by analyzing small, individual pieces of the data.
A desirable property for summarization, which can be obtained with clustering methods, is the geometric interpretability of elements in the summary. For example, in kMeans the elements in the summary are centroids, which are obtained by averaging points in the input data space, and thus are themselves in the same data space, so that  
 one can associate properties (e.g., labels) to the summary points based on the data points these are derived from. 
In kMeans, each point in the dataset can be considered as a $1$-sparse representation based on the nearest cluster center. 
This  leads to hard partitioning of the input data space, which suggests that better summarization is possible if the optimization allows for points to be approximated by a sparse linear combination of summary points. 

In this paper, we investigate data summarization using a dictionary learning~(DL) framework where the summary, or dictionary, is optimized for $k$-sparsity, with $k > 1$, i.e., each data point is represented by a $k$-sparse combination of elements (atoms) from an adaptively learned set, the dictionary. 
It is important to note that using 
% DL has been used successfully in a range of applications, and was originally motivated as a generalization of kMeans, the use of DL for data summarization has not been considered before.  
% The application of 
previous DL schemes such as the method of optimal directions~(MOD) \cite{engan1999method}, the kSVD algorithm \cite{aharon2006k}, and their kernel extensions  \cite{zhang2011kernel, van2012kernel, van2013design}, 
for DL-based data summarization is not possible for several reasons. 
Firstly, current DL methods learn dictionary atoms that are optimized to represent data and 
their \emph{approximation residuals} \cite{dumitrescu2018dictionary}.
This means that atoms in the dictionary are not guaranteed to be points that are on, or even near, the input data manifold and do not have geometric properties as those of cluster centers in  kMeans.
Secondly, 
although DL methods
perform well in signal and image processing tasks, 
their application to machine learning problems is largely limited to learning class-specific dictionaries that can be later used for classification \cite{ramirez2010classification, vu2017fast}. 
This is because the individual atoms learned by DL cannot be directly associated to labels, or other properties of the data, and can only be assigned labels if separate class-wise dictionaries are learned.  
Finally, current DL schemes are impractical even for datasets of modest size \cite{feldman2013learning, golts2016linearized} and are thus not suitable for summarization involving large datasets.

\begin{figure*}[htbp]
    \centering
    \begin{subfigure}{0.37\textwidth}
        \includegraphics[width=\textwidth]{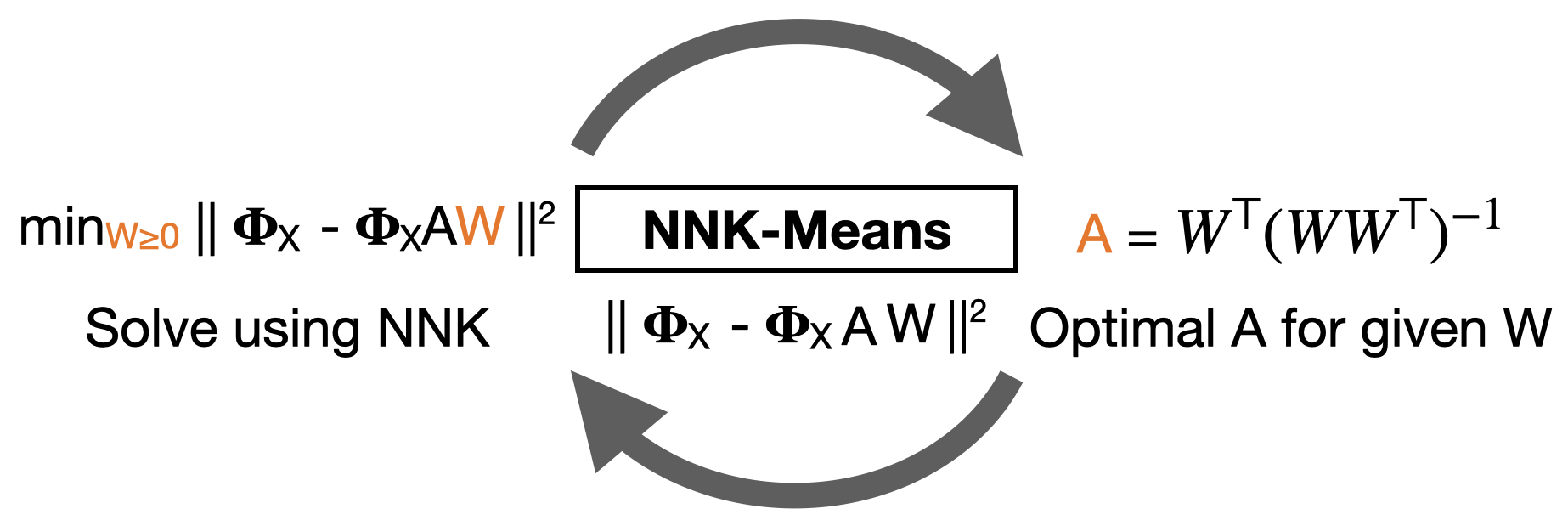}
    \end{subfigure}
    \vrule~
    \begin{subfigure}{0.25\textwidth}
        \includegraphics[width=\textwidth]{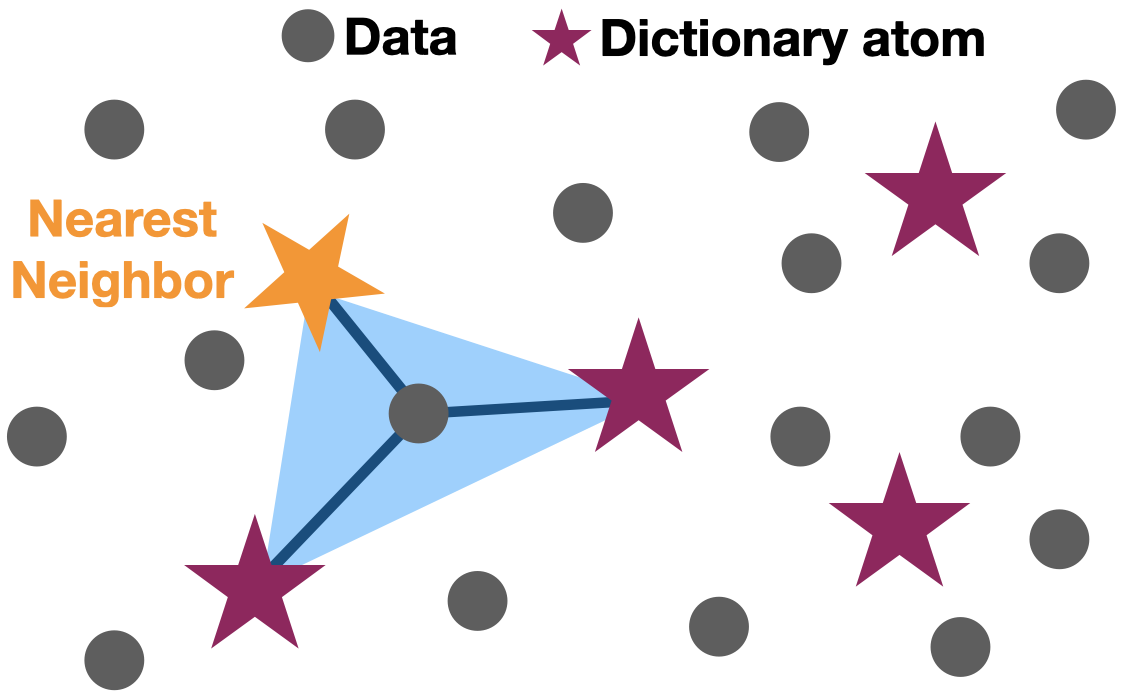}
    \end{subfigure}
    \vrule~
    \begin{subfigure}{0.33\textwidth}
        \includegraphics[width=\textwidth]{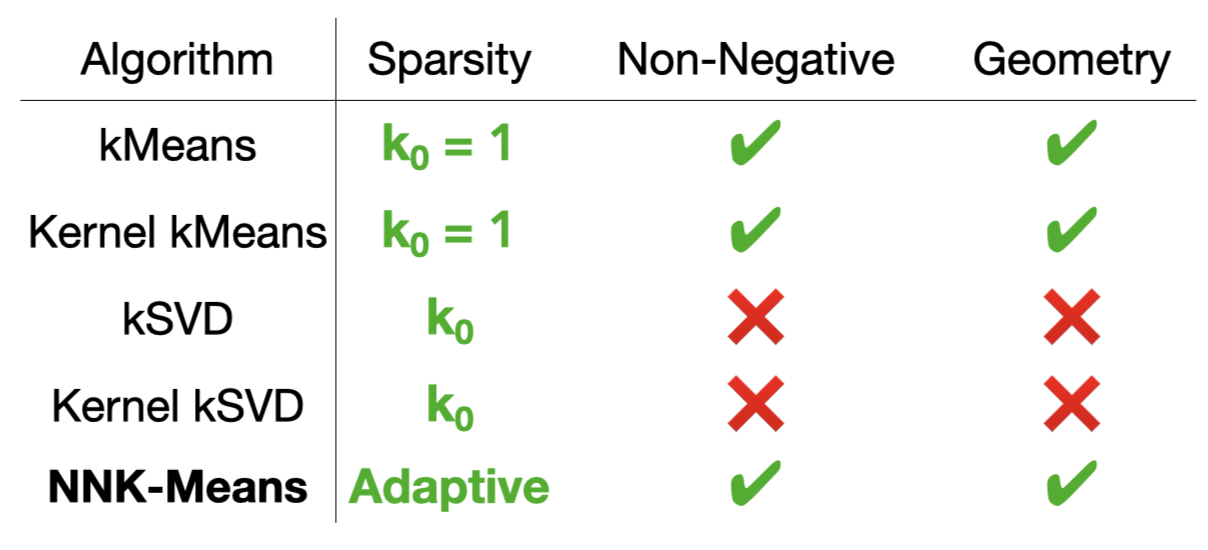}
    \end{subfigure}
    \caption{\textbf{Left:} Proposed NNK-Means. The algorithm alternates between sparse coding ($\mW$) using NNK and dictionary update ($\mA$) until either the dictionary elements converge, or a given number of iterations  or a reconstruction error is achieved.
    \textbf{Middle:} During sparse coding, kMeans assigns each data point to its nearest neighbor while NNK represents each data point in an \emph{adaptively} formed convex polytope made of the dictionary atoms. \textbf{Right:} Comparative summary between dictionary learning methods and proposed NNK-Means approach. kMeans offers a $1$-sparse dictionary learning approach while a kSVD offers a more general approach where the sparse coding stage accommodates for a chosen, fixed $\rk_0$-sparsity but lacks geometry. NNK-Means has adaptive sparsity that relies on the relative positions of atoms around each data to be represented.}
    \label{fig:nnk_means_summary}
\end{figure*}

To overcome these limitations and learn dictionaries with atoms that can be used for data summarization,
we leverage our work on neighborhood definition with non-negative kernel regression (NNK) \cite{shekkizhar2020, shekkizhar2021revisiting}. Our proposed 
\textit{NNK-Means} algorithm for data summarization is based on a dictionary obtained using  
 a sparse coding technique based on NNK, where 
each selected neighbor corresponds to a direction in input space that is not represented by other selected neighbors. 
%, each data point via an \emph{adaptive} set of nearest dictionary atoms selected using NNK 
%(e). 
This representation can be interpreted geometrically as a polytope covering of the data by selected atoms \cite{shekkizhar2020}. 

In all DL methods, as in ours, learning is done by alternating two minimization steps, namely sparse coding and dictionary update, with differences between the methods arising given the choice of constraints or optimization in these two steps. 
The main novelty in NNK-Means comes from the use of a non-negative sparse coding procedure in kernel space that can 
be described, similar to kMeans, in terms  of the local data geometry.
%similar to kMeans. 
Non-negative DL and kernelized DL were separately studied in  \cite{aharon2005k, bevilacqua2013k, pan2014robust} and \cite{van2012kernel, van2013design}, respectively. 
%but neither of these works combined the two approaches.
Closest to our work are  \cite{hosseini2016non, zhang2017image},  where dictionaries are learned in kernel space with non-negative sparse coding performed using optimization schemes, such as an $l_1$ constrained quadratic solver with multiplicative updates \cite{hoyer2002non, zhang2017image, zhou2021kernel} or make use of iterative thresholding algorithms with non-negativity constraint \cite{aharon2005k, bevilacqua2013k, pan2014robust}. In contrast, our framework makes use of a geometric sparse coding approach based on local neighbors, a procedure previously unexplored in DL.  The sparsity of the representation in our approach depends on the relative position of the data and atoms (i.e., the data geometry) and is thus interpretable and adaptive. 
Consequently, unlike earlier DL methods, individual atoms learned by NNK-Means have explicit geometric properties, with representations that are obtained as averages of input data examples similar to kMeans, and can be associated with data properties such as class labels. This makes the atoms learned by our approach suitable for data summarization. Note that previous DL methods and sparse coding approaches lacked such properties, so the proposed NNK-Means is the first DL framework to study these concepts with emphasis on geometry for use in data summarization. 

Our experiments show that the NNK-Means i) selects atoms for summarization that belong to the data space,  
ii) outperforms DL methods in terms of downstream classification using class-specific summaries on several datasets (USPS, MNIST, and CIFAR10), and  
iii) achieves train and test runtimes similar to kernel kMeans, and $67\times$ and $7\times$ faster than kernel kSVD.

\begin{figure*}[htbp]
    \centering
    \begin{subfigure}{0.19\textwidth}
        \includegraphics[width=\textwidth]{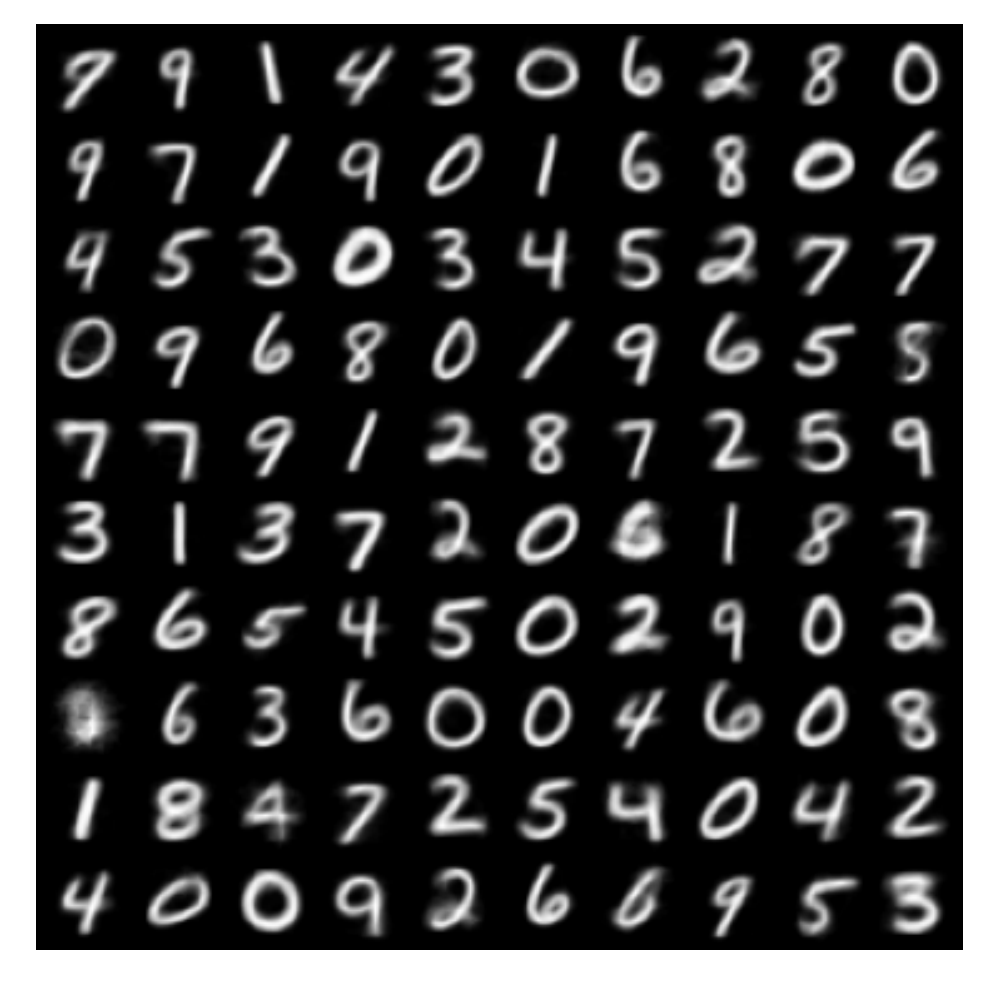}
        \caption{kMeans}
    \end{subfigure}
    % \begin{subfigure}{0.24\textwidth}
    %     \includegraphics[width=\textwidth]{mf_bases/Eigen_PCA.pdf}
    %     \caption{Eigen decomposition (PCA)}
    % \end{subfigure}
    \begin{subfigure}{0.19\textwidth}
        \includegraphics[width=\textwidth]{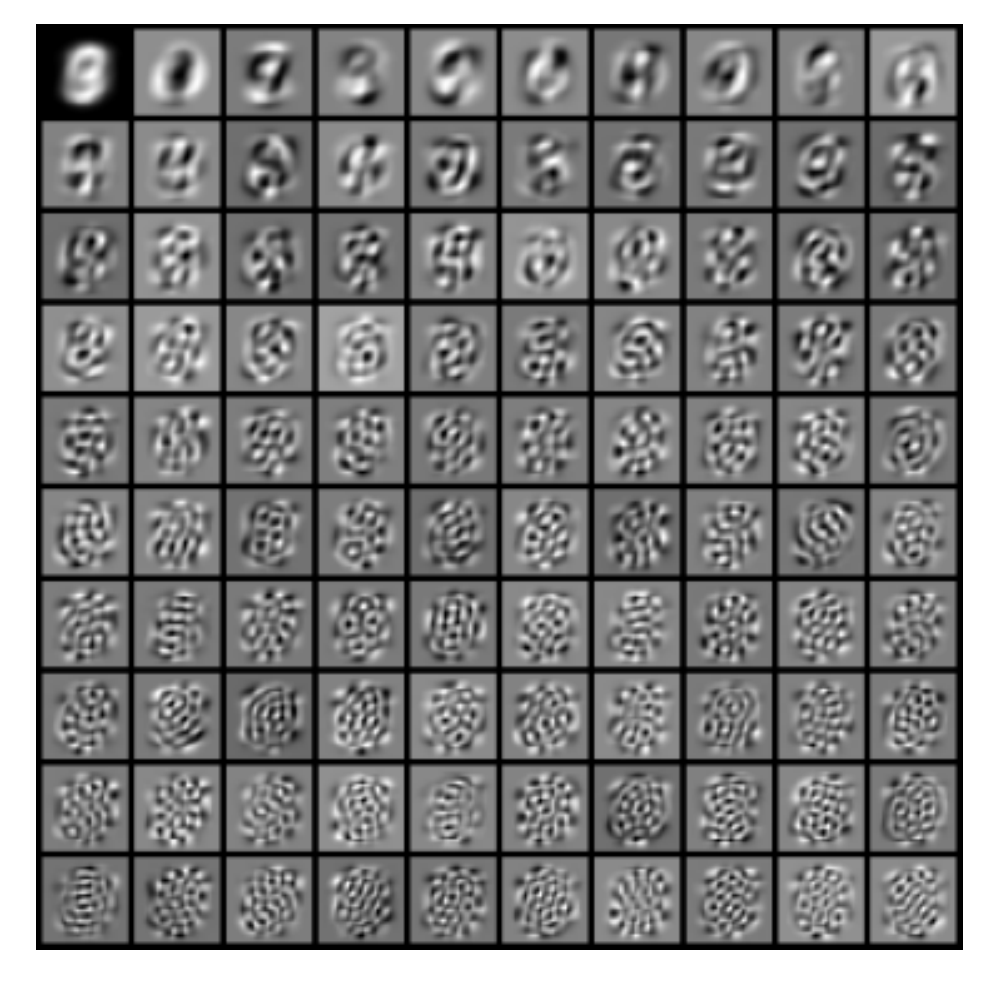}
        \caption{Unconstrained DL}
    \end{subfigure}
    \begin{subfigure}{0.19\textwidth}
        \includegraphics[width=\textwidth]{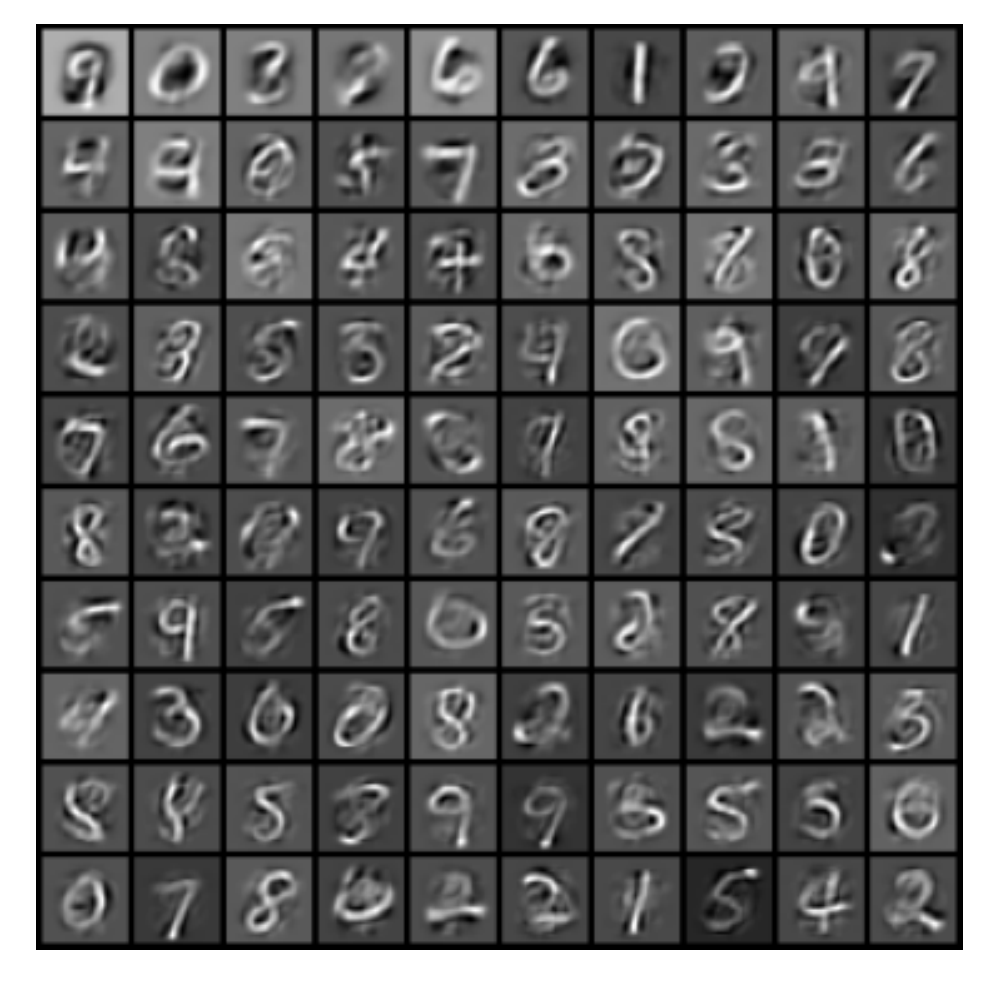}
        \caption{Non-negative $\mW$ ($l_1$ reg.)}
    \end{subfigure}
    \begin{subfigure}{0.19\textwidth}
        \includegraphics[width=\textwidth]{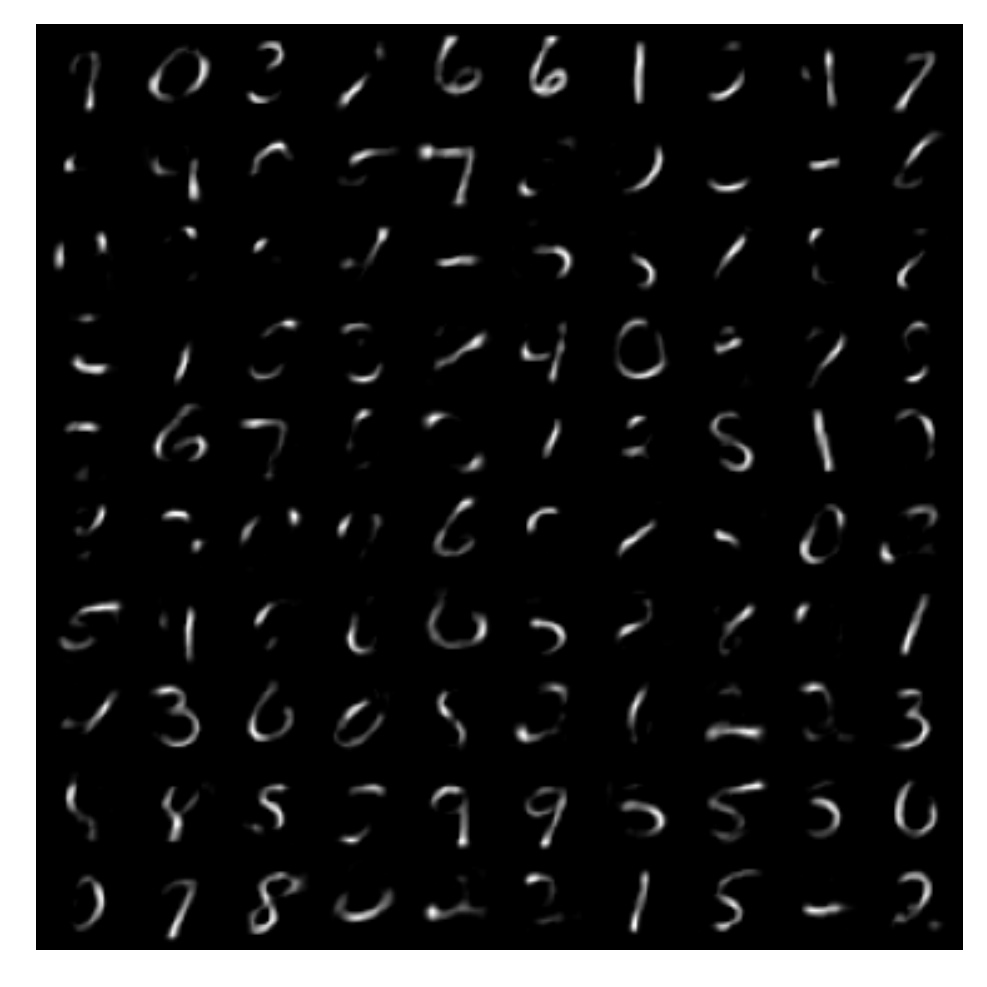}
        \caption{Non-negative $\mD$ and $\mW$}
    \end{subfigure}
    \begin{subfigure}{0.19\textwidth}
        \includegraphics[width=\textwidth]{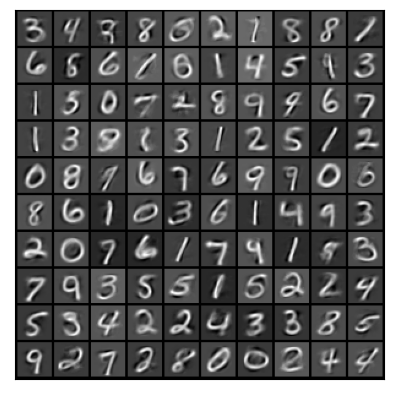}
        \caption{NNK-Means}
    \end{subfigure}
    \caption{$100$ atoms for MNIST digits (contrast scaled for visualization) obtained using kMeans, DL methods and their constrained variants,  NNK-Means (using cosine kernel \cite{shekkizhar2021revisiting}). Unlike earlier DL approaches, our method learns individual atoms that are linear combinations of the digits themselves and can be associated geometrically to the input data. Such explicit properties of atom are lost when working with $l_1$ regularized or thresholding based sparse coding methods used previously in DL.}
    \label{fig:DL_atoms}
\end{figure*}

\section{Problem Setup and Background}
\label{sec:background}
\paragraph*{Sparse Dictionary Learning}
Given a dataset of $N$ data points represented by a matrix $\mX \in \R^{d \times N}$, % $\{\vx_1, \dots \vx_N\}$
the goal of DL is to find a dictionary $\mD \in \R^{d \times M}$ with $M<<N$, and a sparse matrix $\mW \in \R^{M \times N}$ that optimizes data reconstruction.
\begin{align}
    \hat{\mD},\hat{\mW} = \underset{\mD, \mW \colon \forall i \;\; ||\mW_i||_0 \leq \rk }{\argmin} || \mX - \mD\mW ||^2_F \label{eq:dictionary_learning_objective}
\end{align}
where the $\ell_0$ constraint on $\mW$ corresponds to the sparsity requirements on the columns of the reconstruction coefficients $\mW_i \in \R^M$ and $||\;.\;||_F$ represents the Frobenius norm of the reconstruction error associated with the representation. 
% The solution to (\ref{eq:dictionary_learning_objective})  in MOD or kSVD is found in an iterative manner  alternating  between sparse coding ($\mW$) and dictionary update steps ($\mD$). 

While kMeans can be written in terms of the DL objective \plainref{eq:dictionary_learning_objective} with a $1$-sparse constraint on the sparse coding, i.e., each column of $\mW$ can have only one nonzero value, there are several important differences between the two problems.  
In particular we can see that: 
(i) the coefficients involved in the sparse coding of kMeans are non-negative,  
(ii) in kMeans, the sparse coding is based on proximity of the data to the atoms~(i.e., cluster centers), whereas in kSVD or MOD,  coding is done by searching for atoms that maximally correlate with the residual, and (iii) the dictionary updates are different and lead to different dictionaries. 
%However, to tackle problems with nonlinear data and where the downstream goal for learned dictionaries is classification or regression, several modifications to DL have been proposed and studied. 

A straightforward way of kernelizing DL would involve replacing the input data by their respective Reproducing Kernel Hilbert Space~(RKHS) representation. However, such a setup is unable to leverage the \emph{kernel trick} \cite{mercer1909, hofmann2008kernel} and thus to overcome this problem, \cite{van2012kernel} suggest decomposing the dictionary and solving a modified objective \plainref{eq:dictionary_learning_objective}, namely,
% \begin{align}
%     \hat{\mD},\hat{\mW} = \underset{\mD, \mW \colon \forall i \;\; ||\mW_i||_0 \leq \rk }{\argmin} || \mPhi - \mD\mW ||^2_F \label{eq:kernel_dictionary_learning}
% \end{align}
\begin{align}
    \hat{\mA},\hat{\mW} = \underset{\mA, \mW \colon \forall i \;\; ||\mW_i||_0 \leq \rk }{\argmin} || \mPhi - \mPhi\mA\mW ||^2_F \label{eq:reformulated_kernel_dictionary_learning}
\end{align}
where $\mPhi = \phi(\mX)$ corresponds to the RKHS mapping of the  data. In this setup, one learns a dictionary ($\mD = \mPhi\mA$) via the coefficient matrix $\mA \in \R^{N \times M}$.
% This is similar to  kMeans \cite{lloyd1982least, dhillon2004kernel}, although unconstrained, where the atoms are obtained via positive weighted average of input data.
% , namely, $\mA = \mW^\top\mSigma^{-1}$ where $\mSigma \in \R^{M\times M}$ is a diagonal matrix containing the influence of one atom to the data representation. 

\paragraph*{Non-Negative Kernel Regression}
The starting point for our DL method is our graph construction framework using NNK \cite{shekkizhar2020, shekkizhar2021revisiting}. NNK formulates neighborhoods as a signal representation problem, where data points (represented as a RKHS function) is to be approximated by  functions corresponding to its neighbors, i.e., 
% The NNK objective for neighborhood is,
\begin{align}
 \min_{\vtheta \geq 0} \;
 \|\phi(\vx_i) - \mPhi_S\vtheta\|^2, \label{eq:nnk_lle_objective}
\end{align}
where $\mPhi_S$ contains the RKHS representation of a pre-selected set of data points that are good candidates for NNK neighborhood.
Unlike k-nearest neighbor or $\epsilon$-neighborhood, where a neighbor is selected based on only  $\kappa(\vx_i,\vx_j)=\vphi(\vx_i)^\top\vphi(\vx_j)$, and can be viewed as representation using thresholding, NNK leads to optimal neighborhoods, that avoids selecting two neighbors that are similar to each other. 
Geometrically, this can be explained using hyperplanes, one per selected NNK neighbor, which applied inductively leads to a convex polytope around the data such as the one in  \Figref{fig:nnk_means_summary}. 

\section{Proposed Method: NNK-Means}
\label{sec:proposed_method}
% We present a data summarization approach using DL that draws ideas from kMeans and our work on NNK neighborhood \cite{shekkizhar2020}. 
We propose a two-stage learning scheme where we solve sparse coding and dictionary update until convergence, or until a given number of iterations or reconstruction error is reached. We describe the two steps, the respective optimization involved, interpretation, and runtime complexity in this section\footnote{A longer version of the paper is posted on arxiv with proofs of
theoretical statements and additional experiments\cite{shekkizhar2021nnkmeans}}.

\begin{figure*}[htbp]
    \begin{subfigure}{0.16\textwidth}
    \centering
        \includegraphics[trim={1.5cm 1cm 1.5cm 1cm}, clip, width=\textwidth]{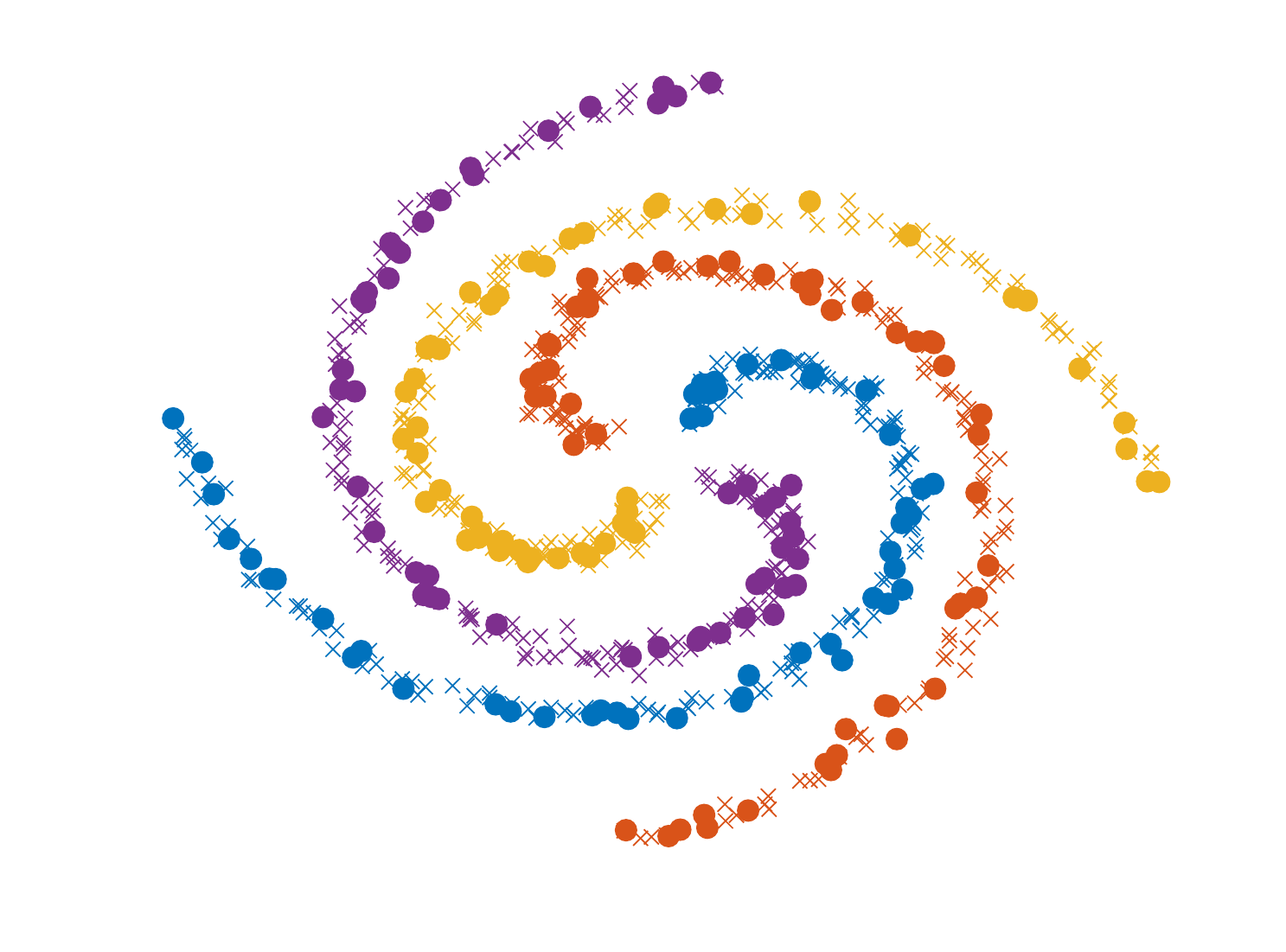}
        \caption{Train \& Test data }
    \end{subfigure}
    \begin{subfigure}{0.16\textwidth}
    \centering
        \includegraphics[trim={1.5cm 1cm 1.5cm 1cm}, clip,width=\textwidth]{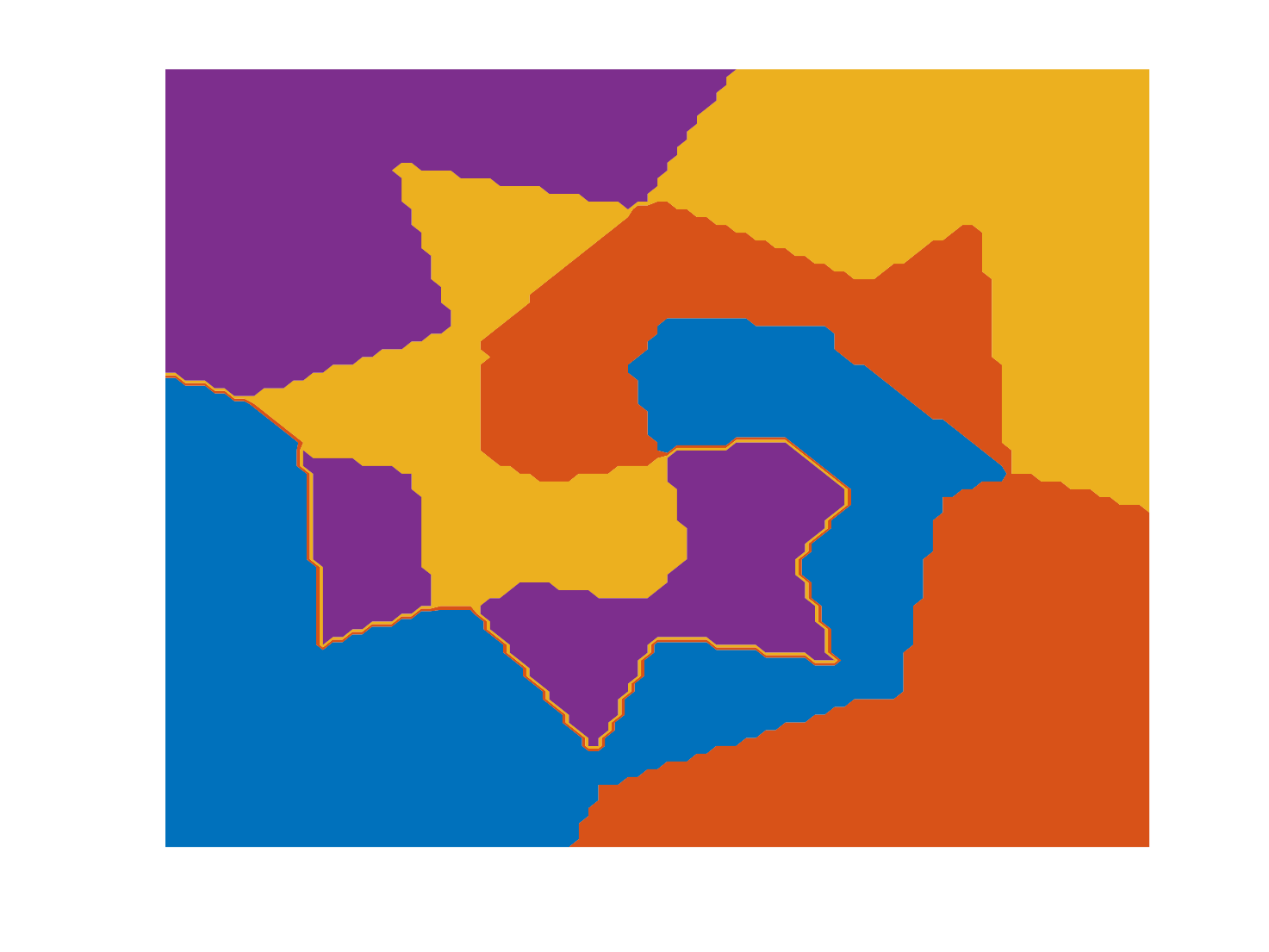}
        \caption{kMeans ($0.85$)}
    \end{subfigure}
    \begin{subfigure}{0.16\textwidth}
    \centering
        \includegraphics[trim={1.5cm 1cm 1.5cm 1cm}, clip,width=\textwidth]{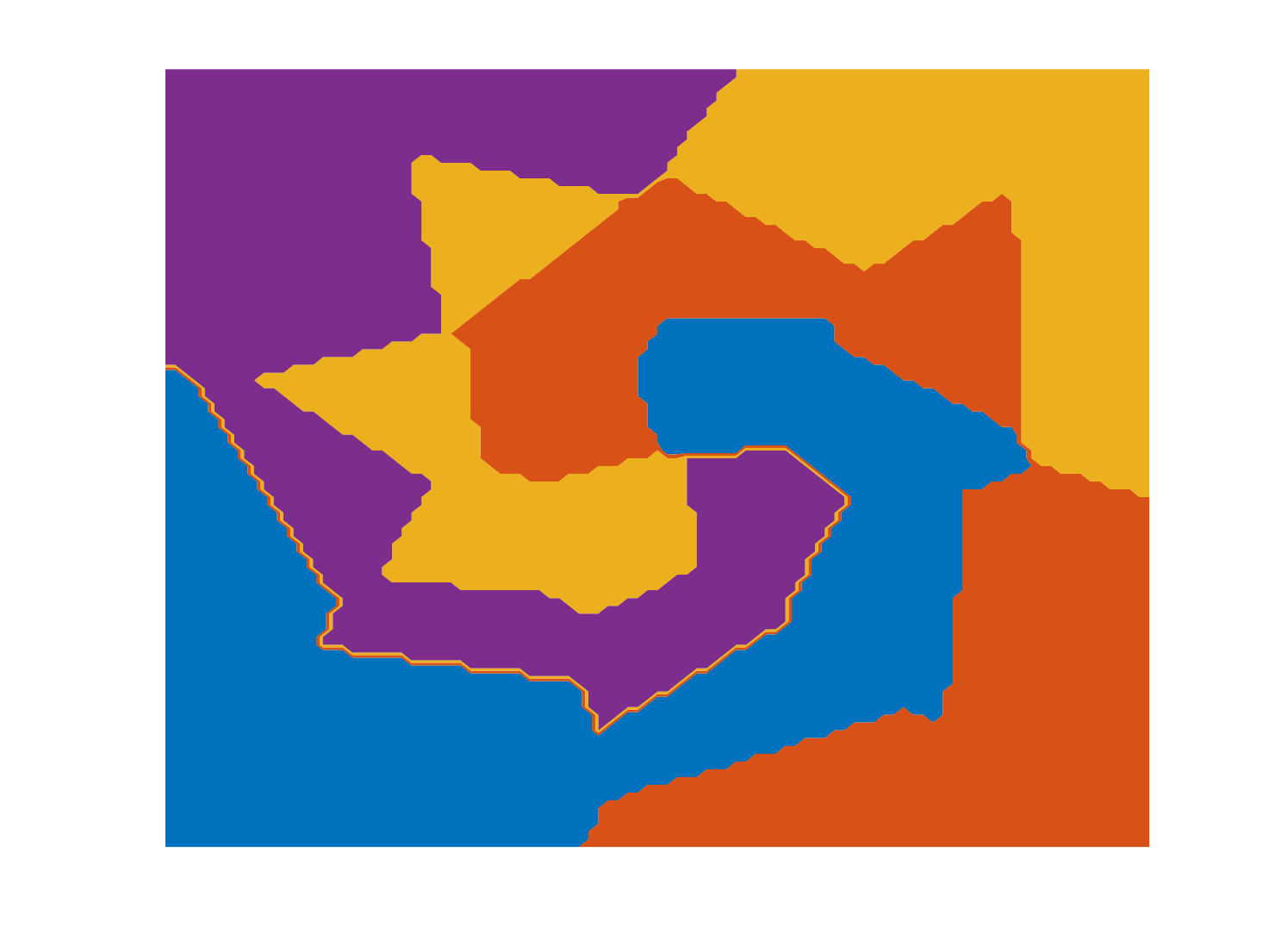}
        \caption{Kernel kMeans ($0.8$)}
    \end{subfigure}
    \begin{subfigure}{0.16\textwidth}
    \centering
        \includegraphics[trim={1.5cm 1cm 1.5cm 1cm}, clip,width=\textwidth]{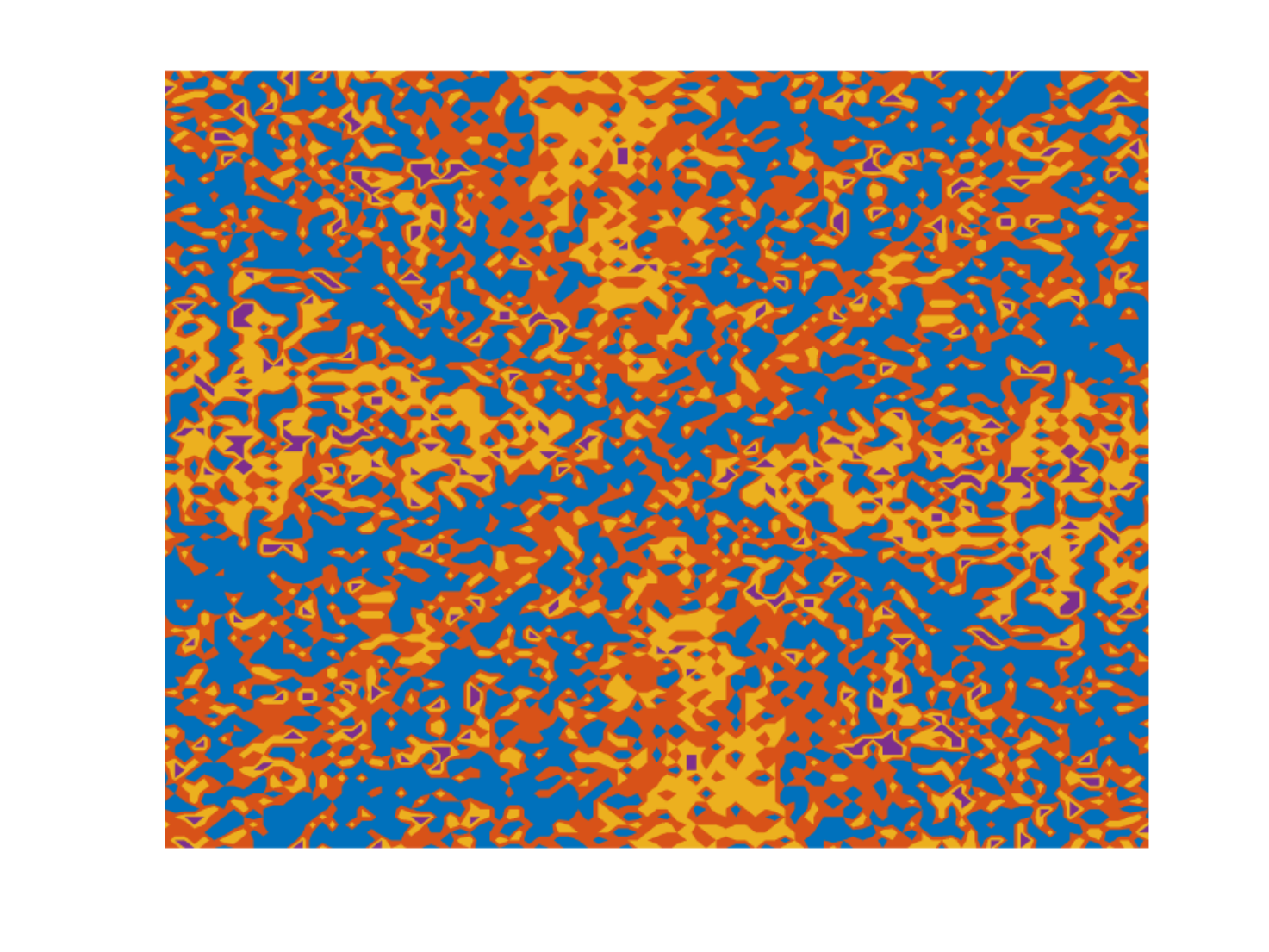}
        \caption{kSVD ($0.24$)}
    \end{subfigure}
    \begin{subfigure}{0.16\textwidth}
    \centering
        \includegraphics[trim={1.5cm 1cm 1.5cm 1cm}, clip,width=\textwidth]{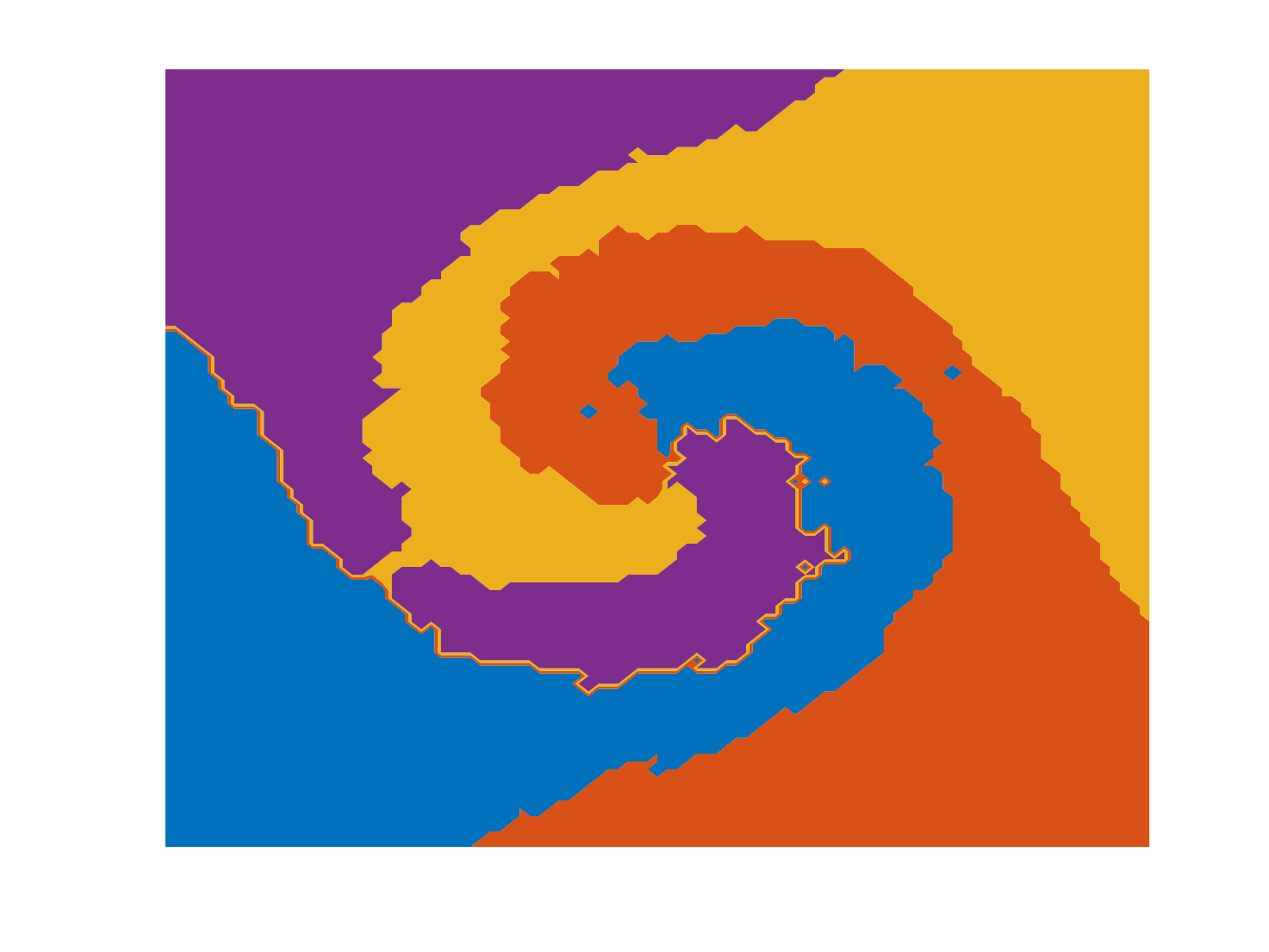}
        \caption{Kernel kSVD ($0.98$)}
    \end{subfigure}
    \begin{subfigure}{0.16\textwidth}
    \centering
        \includegraphics[trim={1.5cm 1cm 1.5cm 1cm}, clip,width=\textwidth]{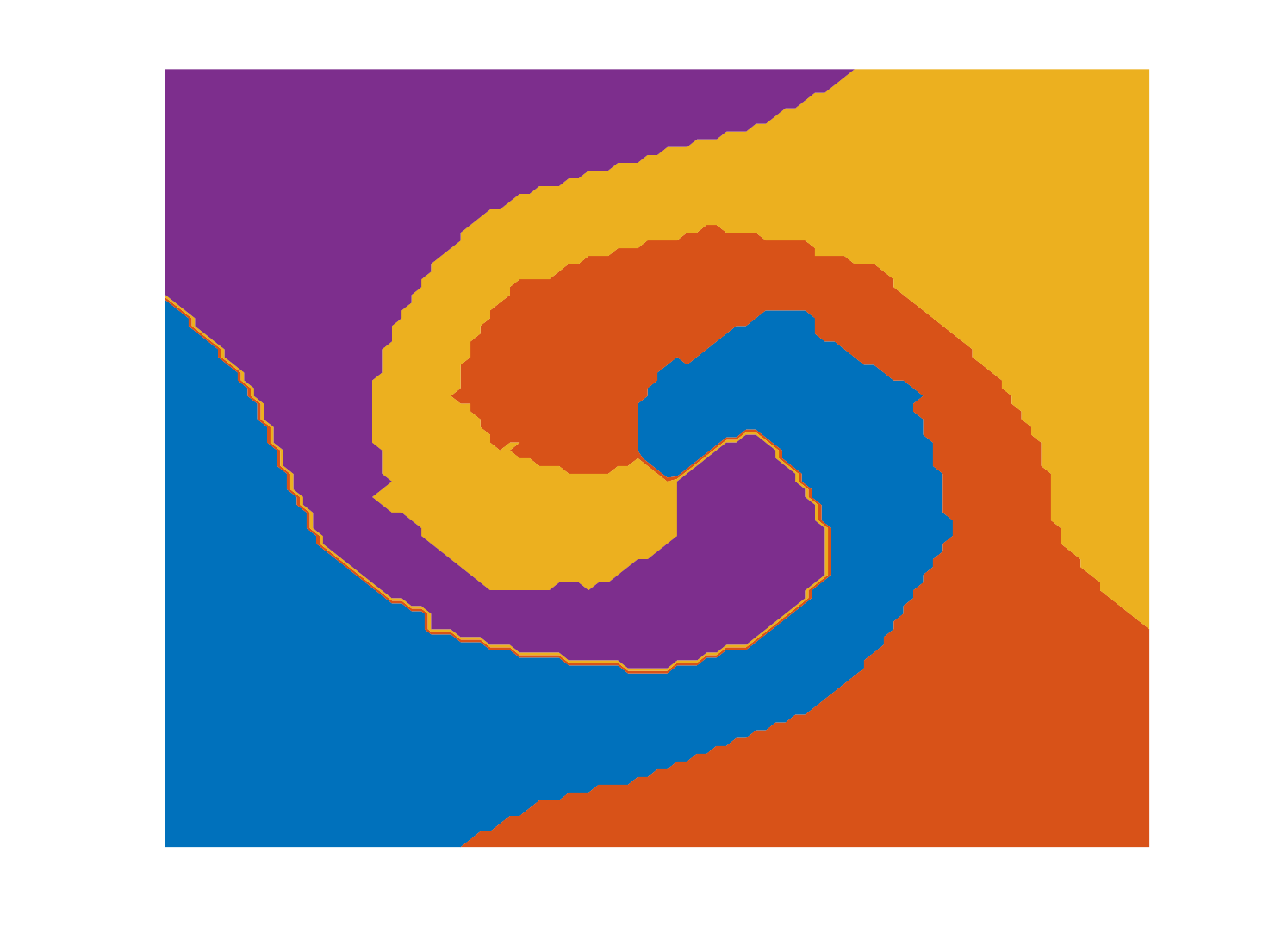}
        \caption{NNK-Means ($1.0$)}
    \end{subfigure}
    \caption{Visualizing predictions (each color corresponds to a class) obtained using various dictionary based classification schemes on a $4$-class synthetic dataset. Each method learns a $10$ atom dictionary per class based on given training data ($N=600$, $\times$) with sparsity constrain, where applicable, of $5$. The learned per class dictionary is then used to classify a test data ($N=200$, $\bullet$) with accuracy as indicated in parenthesis for each method. We see that the kSVD approach is unable to adapt to the nonlinear structure of data and adding a kernel is crucial in such scenarios. NNK-Means better adapts to the geometry of the data with a runtime comparable to that of kMeans while having $4\times$ and $2\times$ faster train and test times in comparison with kernel kSVD.}
    \label{fig:simulated_data_result}
\end{figure*}

\paragraph*{Sparse Coding}
%In this stage, the dictionary representation, here $\mA$, is assumed to be fixed. 
Given a dictionary, $\mA$, in this step we seek to find a sparse matrix $\mW$ that optimizes data reconstruction in kernel space. We will additionally require the coefficients of representation to be non-negative with the number of nonzero coefficients at most $k$. Thus, the objective to minimize at this step is 
\begin{align}
    \hat{\mW}  
    % &=\underset{\forall i \;\; \mW_i \geq 0, \; ||\mW_i||_0 \leq k}{\argmin} || \mPhi - \mPhi\mA\mW ||^2_F  \nonumber\\
    &= \underset{\forall i \;\; \mW_i \geq 0, \; ||\mW_i||_0 \leq k}{\argmin} \sum_{i=1}^N ||\vphi_i - \mPhi\mA\mW_i||^2_{2}, \label{eq:sparse_coding_step}
\end{align}
where $\vphi_i$ corresponds to the RKHS representation of data $\vx_i$. Solving for each $\mW_i$ in \eqref{eq:sparse_coding_step} involves working with a $N\times N$ kernel matrix leading to run times that scale poorly with the size of the dataset. 
However, the geometric understanding of the NNK objective in \cite{shekkizhar2020}, allows us to efficiently solve for the sparse coefficients ($\mW_i$) for each data point by selecting and optimizing starting from a small subset of data points, here the $k$-nearest neighbors. Objective \plainref{eq:sparse_coding_step} can be rewritten for each data point and solved with NNK to obtain $\mW_i$ as 
\begin{align}
        \hat{\mW}_{i, S} &= \underset{\vtheta_i \geq 0}{\argmin} ||\vphi_i - \mPhi\mA_{S}\vtheta_i||^2_{2} 
        \;\; \text{and} \;\; \hat{\mW}_{i, S^c} = \vzero \label{eq:sparse_coding_localized}
\end{align}
where the set $S$ corresponds to the selected subset of indices corresponding to the set of the dictionary atoms $\mPhi\mA$ that can have a nonzero influence in the sparse non-negative reconstruction.
The above reduced objective can be solved efficiently as in NNK graphs \cite{shekkizhar2020}.
% Further, due to the adaptivity of NNK to the relative position of the atoms in the neighborhood of the data, the number of NNK neighbors saturates to a constant and thus one can choose to remove the explicit constrain on the sparsity by choosing a large enough subset $S$.
Sparse coding using NNK allows us to explain the obtained sparse codes, leverage nearest-neighbor tools for scaling to large datasets, and analyze the obtained atoms geometrically, very much similar to kMeans, where each data is represented by an adaptive set of non-redundant neighbors rather than just $1$. This step includes a neighborhood search  and a non-negative quadratic optimization with runtime complexities $\gO(NMd)$ and $\gO(Nk^3)$.

\paragraph*{Dictionary Update}
Assuming that the sparse codes for each training data, $\mW$, are calculated and fixed, the goal  is to update $\mA$ such that the reconstruction error is minimized.
% , i.e.,
% \begin{align}
%     \hat{\mA} = \underset{\mA}{\argmin} || \mPhi - \mPhi\mA\mW ||^2_F \label{eq:dictionary_update_step}
% \end{align}
Here, we propose an update similar to MOD, where the dictionary  matrix $\mA$ is obtained based on $\mW$ as
\begin{align}
    \hat{\mA} = \mW^\top(\mW\mW^\top)^{-1} \label{eq:dictionary_update}
\end{align}
The runtime associated with this step is $\gO(M^3 + NMk)$, where we use the fact that $\mW$ has at most $Nk$ non zero elements. We note that using $k$-nearest neighbor directly for sparse coding, apart from lacking adaptivity, is sub-optimal and leads to instabilities at the dictionary update stage and thus is unsuitable for DL in a similar setup.
\begin{proposition}
\label{prop:relation_to_k_means}
The dictionary update rule in \plainref{eq:dictionary_update} reduces to kMeans cluster update $\mA = \mW^\top\mSigma^{-1}$  when $\mW$ consists of $N$ columns from $(\ve_1 \dots \ve_M)$, where $\ve_m$ is a basis vector, i.e., 
$
    e_{mi} = 0 \; \forall \;i\neq m \text{ and }e_{mm} = 1 
$
and $\mSigma \in \R^{M\times M}$ is a diagonal matrix containing the degree or number of times each basis vector $\ve_m$ appears in $\mW$.
\end{proposition}

Proposition \ref{prop:relation_to_k_means}
shows that our proposed method reduces to the kMeans algorithm when the sparsity of each column in $\mW$ is constrained to $1$ and can thus be considered a DL generalization that maintains the geometric and interpretable properties of kMeans.
One can easily verify that our iterative procedure for DL, alternating between sparse coding and dictionary update, does converge (\autoref{thm:convergence}) and produces atoms that belong to the input data manifold. 
% and Proposition \ref{prop:nonnegative_projecion_matrix} respectively. 
% The pseudo-code for NNK-Means is presented in \autoref{tab:nnk_means_algorithm}.

\begin{theorem}
\label{thm:convergence}
The residual $|| \mPhi - \mPhi\mA\mW ||^2_F$ decreases monotonically under the NNK sparse coding step (\ref{eq:sparse_coding_localized}) for $\mW$  given matrix $\mA$ . For a fixed $\mW$, the dictionary update  (\ref{eq:dictionary_update}) for $\mA$ is the optimal solution to $\min_{\mA}|| \mPhi - \mPhi\mA\mW ||^2_F$ . Thus, NNK-Means objective decreases monotonically  and converges.
\end{theorem}

\section{Experiments}
\label{sec:experiments}
In this section, we validate properties of NNK-Means that make it suitable for data summarization.
\Figref{fig:DL_atoms} presents a visual comparison of the atoms obtained using our method with that of kMeans and previous DL approaches. Unlike standard DL approaches, we observe that atoms learned by NNK-Means have representations that are similar to the input data.
We will now focus on a standard experiment setting in DL, namely DL-based classification \cite{van2012kernel, golts2016linearized}, to compare NNK-Means with previous DL approaches.
Note that learning a good summary leads to better classification.  
Since existing DL methods cannot associate labels directly to the atoms obtained, experiments are constrained to learning a dictionary for each class ($\{\mA_i\}^{C}_{i=1}$) in training data that are later used to classify queries based on the class-specific reconstruction error $\{e_i\}^{C}_{i=1}$, i.e., we sparse code a query $\vx_q$ using each dictionary $\mA_i$ and assign the query to the class~($c$) with lowest reconstruction error~($e_c$). 
NNK-Means outperforms all other methods consistently in classification while having desirable runtimes relative to kMeans, kSVD, and their kernelized versions\footnote{We use the efficient implementations,  as in \cite{golts2016linearized}, from omp-box and kSVD-box libraries \cite{rubinstein2008efficient} and Kernel kSVD code of \cite{van2012kernel}.} in both synthetic and real datasets.   
We use a Gaussian kernel $\kappa(\vx, \vy) = \text{exp}(||\vx - \vy ||_2^2/2)$  and report average performance over $10$ runs for all experiments. 
% \footnote{Source code  available at \href{https://github.com/STAC-USC/NNK\_Means}{github.com/STAC-USC/NNK\_Means}}

\paragraph*{Synthetic dataset} 
We consider a $4$-class dataset consisting of samples generated from a non-linear manifold and corrupted with gaussian noise (as in \Figref{fig:simulated_data_result}). 
% The first plot in \Figref{fig:simulated_data_result} shows the training data and test data used for the experiment with their corresponding labels. 
Since the data corresponding to each class have similar support, namely the entire space $\R^2$, dictionaries learned using kSVD are indistinguishable for each class and lead to at-chance performance in classification of test queries. On the contrary, a kernelized version of kSVD is able to handle the manifold, although it is not robust at some test locations, but at the cost of increased computational complexity. Interestingly, we observe that a non-negative neighborhood-based sparse coding is able to adapt to input space non-linearity even when constrained to $1$-sparsity (kMeans) indicative of the importance of non-negativity and geometry in data summarization. 

\begin{figure*}
\centering
% \begin{subtable}{0.22\textwidth}
% \centering
%     \begin{tabular}{l c c c }
%     \toprule
%     Dataset & $d$ & $N_{train}$ & $N_{test}$\\\midrule
%     USPS & $256$ & $7291$ & $2007$ \\
%     MNIST & $784$ & $60000$ & $10000$ \\
%     MNIST-S & $784$ & $12000$ & $10000$ \\
%     CIFAR  & $512$ & $50000$ & $10000$ \\
%     CIFAR-S  & $512$ & $10000$ & $10000$ \\\bottomrule
%     \end{tabular}
% \end{subtable}
% ~
\begin{subfigure}{0.3\textwidth}
\includegraphics[trim={0cm 0cm 1cm 0.4cm},clip,width=\textwidth]{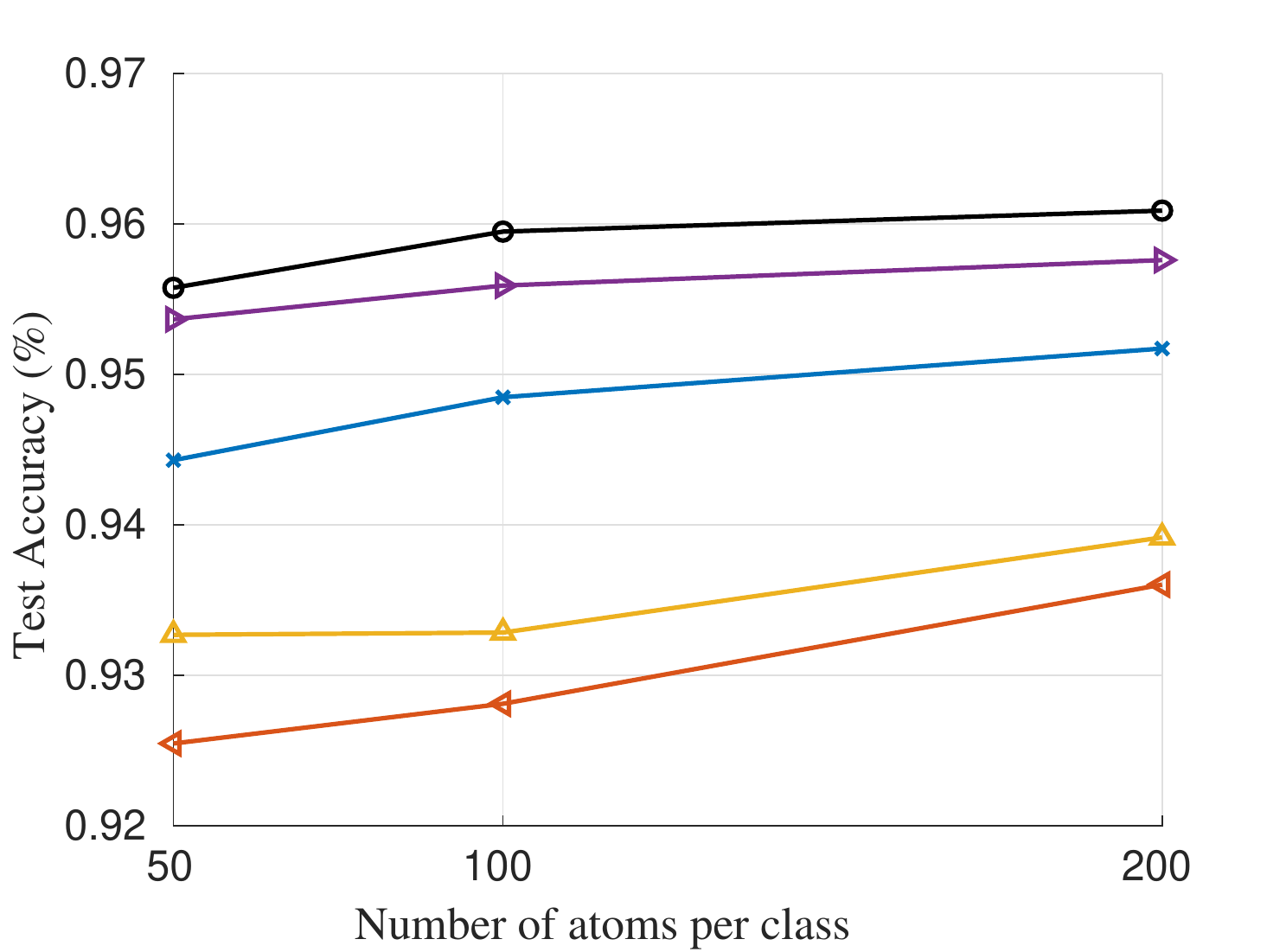}
\end{subfigure}
\begin{subfigure}{0.3\textwidth}
\includegraphics[trim={0cm 0cm 1cm 0.4cm},clip,width=\textwidth]{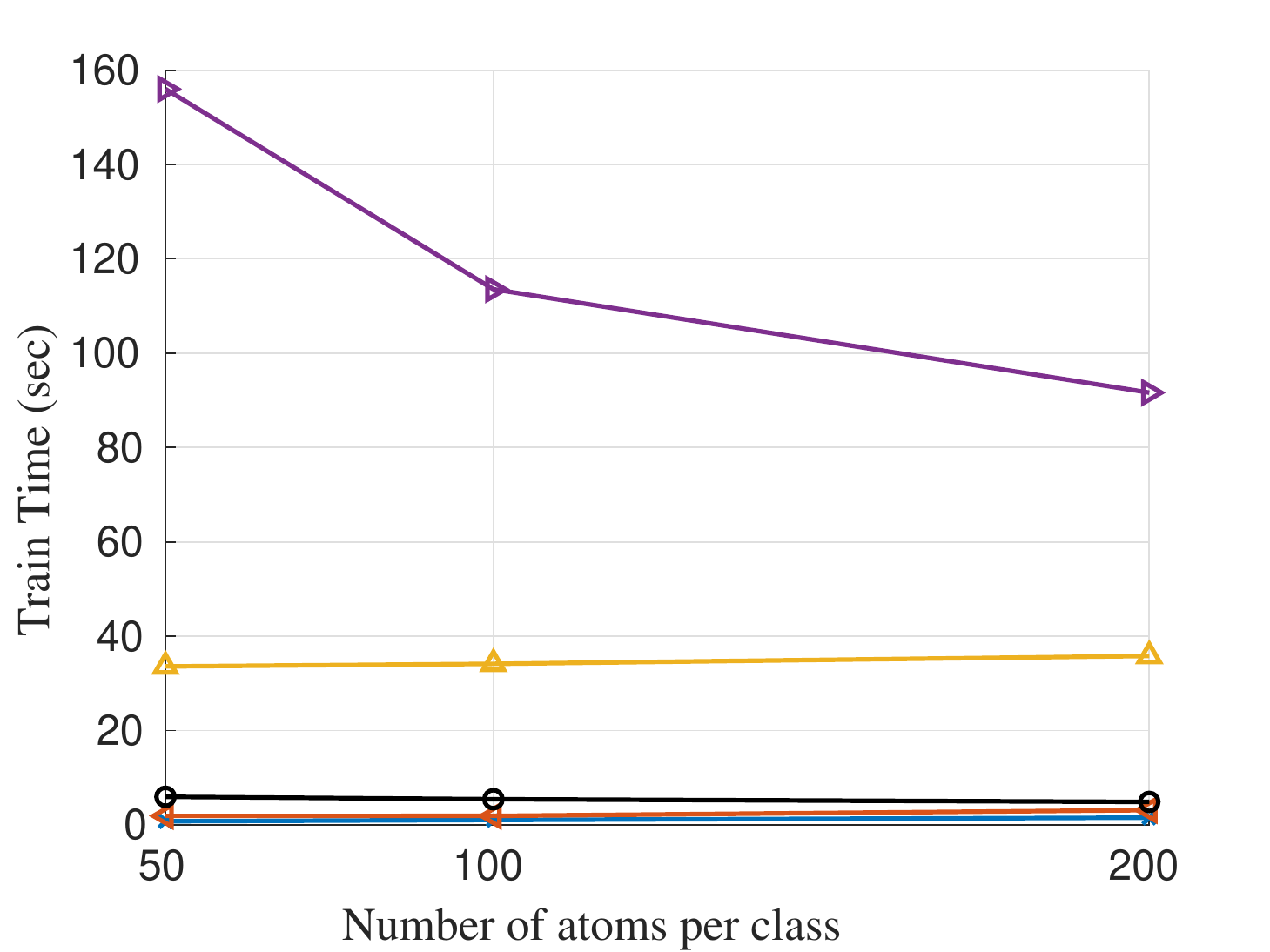}
\end{subfigure}
\begin{subfigure}{0.3\textwidth}
\includegraphics[trim={0cm 0cm 1cm 0.4cm},clip,width=\textwidth]{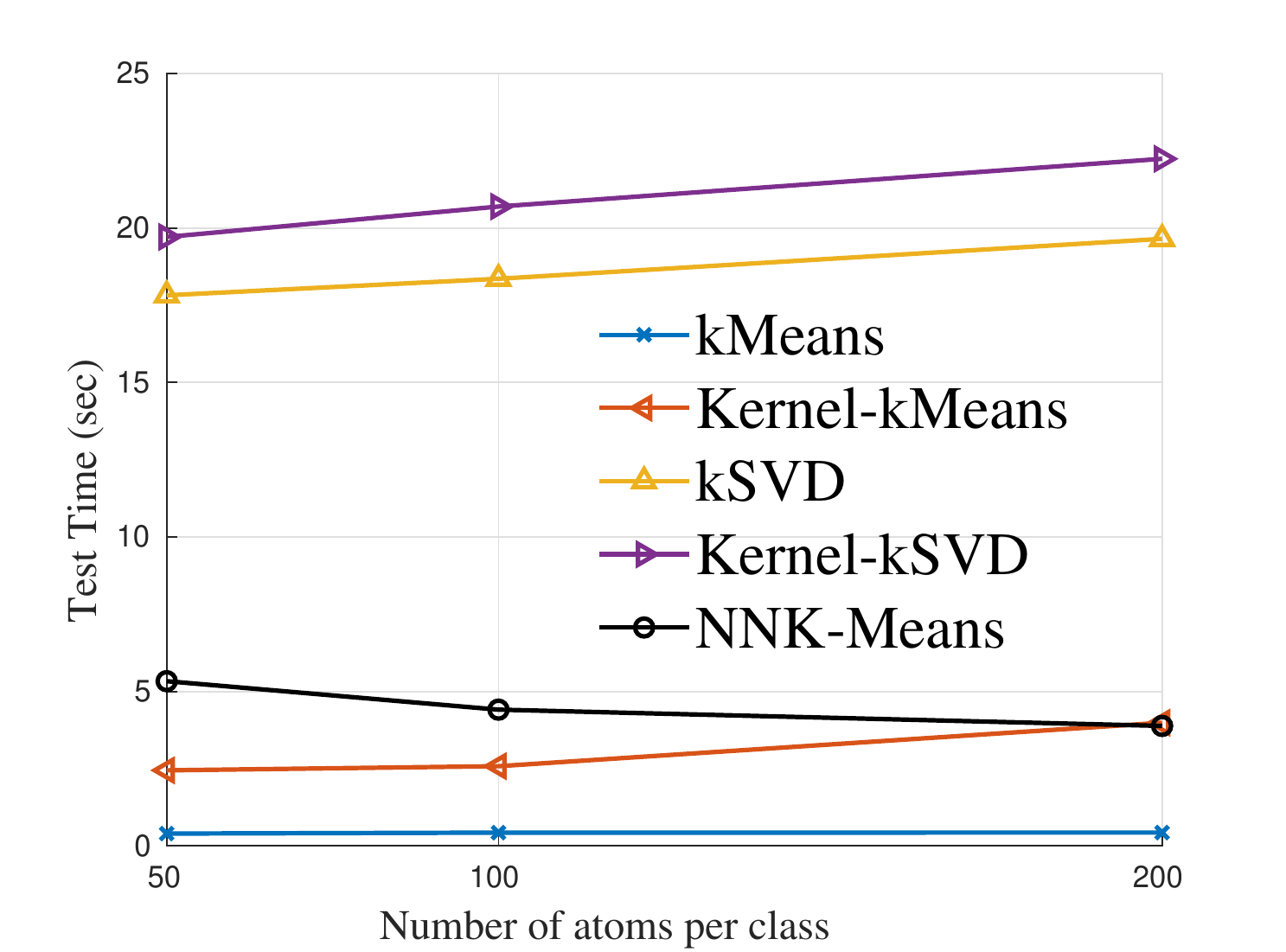}
\end{subfigure}
\caption{
% \textbf{Table:} Summary of datasets (dimensionality $d$, number of training and test data points) used for evaluation in experiments. 
\textbf{Plots:} 
Test classification accuracy, train time and test time as a function of number of dictionary atoms per class on USPS dataset for various DL methods. Each method is initialized similarly and is trained for a maximum of $10$ iterations with sparsity constraint, where applicable, of $30$. The plots demonstrate the benefits of NNK-Means both in accuracy of classification and runtime. The major gain in runtime for NNK-Means comes from the pre-selection of atoms in the form of nearest neighbors which leads to fast sparse coding (as can be seen via test time which performs only sparse coding) relative to kSVD approaches that sequentially perform a linear search for atoms that correlate with the residue at that stage. The training time in kSVD approaches decreases with more atoms since the sparse coding stage requires fewer atom selection steps compared to those where the dictionary is small. 
Source code for all experiments are made available at \href{https://github.com/STAC-USC/}{github.com/STAC-USC}.
}
\label{fig:usps_dataset}
\end{figure*}

\begin{table}[htbp]
    \centering
    \begin{tabular}{l c c c c}
    \toprule
     Method & MNIST-S & MNIST & CIFAR-S & CIFAR  \\\midrule
     kMeans & $94.89$ & $96.34$ & $83.88$ & $84.91$\\
     K-kMeans & $91.56$ & $93.19$& $84.22$ & $85.06$\\
     kSVD & $95.53$ & $95.86$& $86.01$ & $86.28$\\
     K-kSVD & $96.45$ & $-$ & $86.71$ & $-$\\
     NNK-Means  & $\mathbf{96.70}$ & $\mathbf{97.79}$ & $\mathbf{86.95}$ & $\mathbf{87.21}$\\\bottomrule
    \end{tabular}
    \caption{Classification accuracy (in $\%$, higher is better) on MNIST, CIFAR10 and their subset (S, $20\%$ of randomly sampled training set). Each method learns a $50$-atom dictionary per class, initialized randomly, with  sparsity constraint, where applicable, of $30$ and run for at most $10$ iterations. NNK-Means consistently produces better classification in terms of test accuracy while having a reduced runtime in comparison to kSVD approaches and comparable to that of kMeans. Kernel k-SVD produces comparable performance but at the cost of $67\times$ and $7\times$ slower train and test time relative to NNK-Means.}
    \label{tab:dataset_results}
\end{table}

\paragraph*{USPS, MNIST, CIFAR10} 
% We demonstrate the performance of our method to summarize high dimensional real data with USPS, MNIST and CIFAR10 datasets. 
We use as features the pixel values of the images for USPS~($d=256$) and MNIST~($d=784$) dataset. For CIFAR10, we train a self-supervised model using SimCLR loss \cite{chen2020simple} on unlabelled training data to obtain features~($d=512$) for our experiment.
We use the standard train/test split for each dataset and standardize the feature vectors to zero mean and unit variance.
We report here results of DL with a subset of the training data, namely MNIST-S and CIFAR10-S, for a fair comparison with kernel kSVD. We note that kernel kSVD scaled poorly with dataset size and timed out when working with the entire training set of MNIST and CIFAR10. 
% Reported results are the average over $10$ runs.
% \autoref{tab:dataset_results} shows classification accuracy of various methods on presented datasets. 
NNK-Means is able to efficiently learn a compact set of atoms that are capable of representing each class which in turn provides better classification of test data in all settings as made evident in  \Figref{fig:usps_dataset} and the results in \autoref{tab:dataset_results}. 

% \paragraph*{Discussion:}

\section{Conclusion}
\label{sec:conclusion}
% We have presented a geometric dictionary learning framework for data summarization with an efficient and scalable algorithm for learning and working with the obtained summaries.
We investigate data summarization using DL and propose a framework, NNK-Means, that overcomes the limitations of previous DL methods for summarization.   NNK-Means learns atoms that are geometric like kMeans centroids and 
% We present a scalable, kernelized dictionary learning approach for data summarization with sparsity constraints and an appropriate kernel function that measures similarity in high dimensional data. 
leverages neighborhood tools to efficiently perform sparse coding and adaptively represent data using learned summary elements or atoms of the dictionary. 
Experiments show that our method has runtimes similar to kMeans while learning dictionaries that can provide better discrimination than competing methods.
% NNK-Means outperforms dictionary representation based classification methods in real world datasets with runtimes comparable to that of kMeans. 
% We plan to study NNK-Means algorithm to better position the gains in our approach. We believe that the geometry of atoms obtained with NNK-Means allows for better interpretability and 
In the future, we plan to study the trade-offs associated with summary size and the use of obtained summaries in improving analysis and design of data-driven machine learning systems. 

% Further, as part of this investigation we plan to expand on the connections of our framework to non-negative matrix factorization schemes and study the usefulness of our method in a variety of applied settings.
% \pagebreak

\bibliographystyle{ieeetr}
\bibliography{root_eusipco}

\end{document}